# Automated Tumor Segmentation and Brain Mapping for the Tumor Area

Pranay Manocha[*1], Snehal Bhasme[*1], Tanvi Gupta[2], B.K Panigrahi[2],

Tapan K. Gandhi[2]

[1] Indian Institute of Technology- Guwahati, India 781039
{m.pranay, b.walmik}@iitg.ernet.in
[2] Indian Institute of Technology-Delhi, India 110016
{tan0036@gmail.com, bkpanigrahi@ee.iitd.ac.in,
tgandhi@ee.iitd.ac.in}

**Abstract.** Magnetic Resonance Imaging (MRI) is an important diagnostic tool for precise detection of various pathologies. Magnetic Resonance (MR) is more preferred than Computed Tomography (CT) due to the high resolution in MR images which help in better detection of neurological conditions. Graphical user interface (GUI) aided disease detection has become increasingly useful due to the increasing workload of doctors. In this proposed work, a novel two steps GUI technique for brain tumor segmentation as well as Brodmann area detection of the segmented tumor is proposed. A data set of T2 weighted images of 15 patients is used for validating the proposed method. The patient data incorporates variations in ethnicities, gender (male and female) and age (25-50), thus enhancing the authenticity of the proposed method.
The tumors were segmented using Fuzzy C Means Clustering and Brodmann area detection was done using a known template, mapping each area to the segmented tumor image. The proposed method was found to be fairly accurate and robust in detecting tumor.

## 1  Introduction

Diagnosing neurological disorders is a challenging domain. This is because of highly complex anatomical structure of the brain along with some other factors like variations in size, age, previous history, gender, ethnicity and most importantly variations of physiology and also pathology. This requires a comprehensive clinical diagnosis of the patient which may require multiple sequences to be run. Manual detection by doctors is time consuming and prone to human errors and hence automatic detection techniques could aide in providing fast and accurate healthcare.

Tumors are an abnormal growth which can vary in size, location and type. The patient may have a single or multiple tumors at different locations in the brain. MRI (Magnetic Resonance Imaging) is one of the most advanced technology for diagnosing neurological abnormalities like tumors. It is preferred over CT Scan, X-Ray and Ultrasound for obtaining high-resolution images. There are several MR sequences like T1, T1 contrast, T2, FLAIR, PD which are used for detection of various disorders like stroke, cysts, tumor as well as diagnosing of various neurodevelopmental disorders like Parkinson's and Alzheimer's. As each patient data consists of a large number of sequences which requires time to analyses, therefore, automation would aide diagnosticians in reducing the time taken to diagnose the patient as well as make a more accurate diagnosis.

In the past, various methods have been proposed for tumor segmentation. Kraus et al. developed an algorithm on automatic template based segmentation, which has used brain atlas for statistical classification process that divided the image into different parts based on the signal intensity value [1]. The automatic segmentation accuracy varied only 0.6% compared to the manual segmentation accuracy. The algorithm was tested for 20 patients. An atlas is used for the segmentation process and is therefore dependent on an existing data base. Corso et al. developed a brain tumor segmentation technique using Bayesian formulation for calculating affinities & using it in Multi-level segmentation by weighted aggregation algorithm. They have used T1 contrast, T2 and FLAIR sequences for 20 patient data. The accuracy was found to be 70% [2]. The segmentation is done but the area in which the tumor appears is not identified by name. Moon et al. used the expectation maximization (EM) algorithm to modify the probabilistic brain atlas using the spatial priors of the tumor which are obtained using T1 weighted & T1 post contrast difference images. The dataset consisted of T1, T2, T1 post-contrast MR images [3]. Pereira et al. used a 3X3 patch kernel for training convolutional neural network (CNN). Bias field correction, patch and intensity normalization were used for pre-processing. Stochastic Gradient Descent is used to minimize the CNN loss function. The idea of data augmentation in brain tumor segmentation was also explored in that work. For testing they used the BRATS 2013 & 2015 databases. Each patient had T1, T1 contrast, T2 and FLAIR MR sequences [4].

Vergleichende et al. proposed a method where the cerebral cortex was divided in 47 distinct regions as visible in the cell-body stained histological sections. This cyto-architectonic map of the human brain was revolutionary in itself as it helped doctors understand the human brain like never before [5]. Amunts et al. suggested a 3 level concept which included repetitive modular structure and micro and miso maps [6]. Fischl et al. proposed a new way to label the brain areas. The method used the Bayesian framework to label the brain as it allowed explicit incorporation of prior information [7]. Thompson et al. created a probability space of random transformations, based on the theory of anisotropic Gaussian Random Fields to represent the probabilistic brain atlas with an aim to detect any changes in the normal anatomy of the human brain [8]. Shattuck et al. proposed a new method for creating a probabilistic brain atlas from a set of T1-weighted MRI volumes of 40 patients. The atlas was constructed from manually delineated MRI data. Delineation was performed using the trained rafters into 56 structures. Then 3 normalization algorithms were used to create 3 brain templates. Average calculated at each voxel location was used to estimate the probability of that voxel belonging to one of the 56 structures [9].

In this paper, a novel GUI based tumor segmentation and probable mapping to its corresponding Brodmann area is proposed. This proposed algorithm is tested on a dataset of T2 sequences of 15 patients, taking into consideration all possible variations in age, gender and ethnicity. This paper is organized as follows: Section 2 discusses the methodology used for tumor segmentation and GUI development for Brodmann area identification. Section 3 analyses and discusses the results obtained. Section 4 concludes the work done.

## 2 Methodology

Our work has been divided into two sections to comprehensively segment tumor and find the corresponding Brodmann areas and anatomical description of the tumor regions. The first section segments the tumor by forming a neat boundary over the tumor and the second section uses this segmented image and displays the Brodmann areas of the brain in which the tumor exists. The proposed method is depicted in Fig 1.

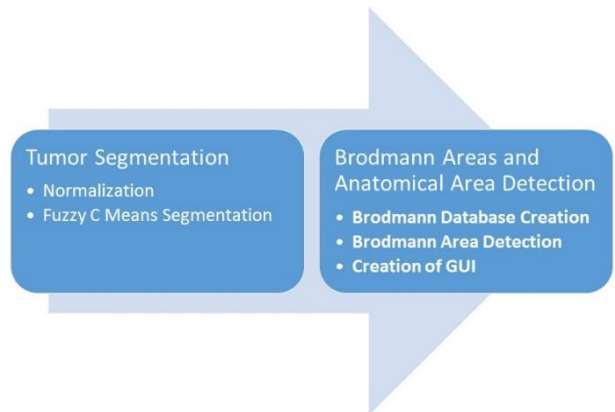

**Fig. 1.** Proposed Method

### 2.1 Tumor Segmentation

**Normalization**

The patient data was normalized using T2.nii template on the SPM toolbox in MATLAB. A 10mm gap between the slices was taken and so 16 slices per patient were obtained.

**Fuzzy C Means**

Fuzzy clustering or soft clustering is a clustering technique in which each data point can belong to more than one clusters. This is in comparison to K means clustering technique or hard clustering which assigns a data point to a single cluster.

Clustering is a method of assigning group or cluster to the different data points, such that items in the same class or cluster are as similar as possible, while items belonging to different classes are as dissimilar as possible.

The steps for Fuzzy Clustering are as follows:

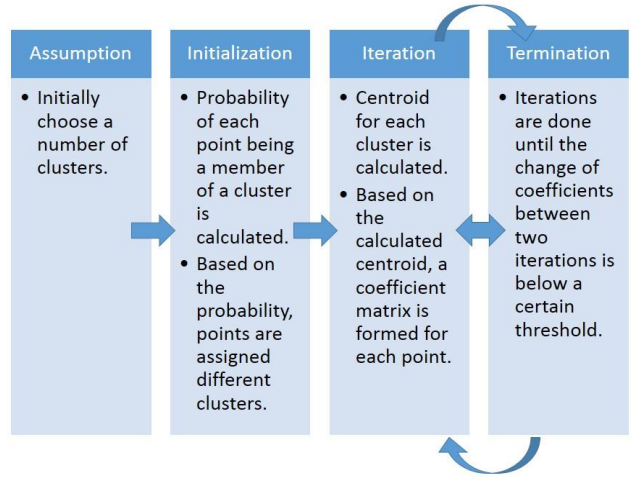

**Fig. 2.** Fuzzy Clustering Method

- **Assumption**: Initially choose a number of clusters, c.
- **Initialization**: Randomly assign each point a cluster coefficient, compute the probability that each data point xi is a member of a given cluster k.
- **Iteration**: Repeat until the algorithm has converged by:

— Computing the centroid for each cluster.

$$C_k = \frac{\sum_x w_k(x)^m x}{\sum_x w_k(x)^m} \quad (1)$$

— Computing the coefficient matrix is for each point.

$$argmin_C \sum_{i=1}^{n} \sum_{j=1}^{c} w_{ij}{}^n \|x_i - c_j\|^2 \quad (2)$$

$$w_{ij} = \frac{1}{\sum_{k=1}^{c}\left(\left\|\frac{x_i - c_j}{x_i - c_k}\right\|\right)^{\frac{2}{m-1}}} \tag{3}$$

- **Termination**: Iteration is carried out until the difference in coefficients between two iterations is no more than a given sensitivity threshold.

### 2.2 Brodmann Area and Anatomical Brain Area Detection

**Brodmann Database creation.**

A database consisting of 47 Brodmann areas was organized for right and left hemispheres. For each hemisphere a database of different Brodmann areas was created for each slice by masking each Brodmann area using a known template.

**Brodmann Area Detection.**

A 3D matrix using the above database of Brodmann areas was created with dimensions 79x95xn where n is the number of Brodmann Areas in a particular slice which is fixed for each slice for each hemisphere.

The matrix addition of the binary segmented tumor image and Brodmann Area images was done and the coordinates whose intensity exceeded '1' were taken into consideration and the slice number was noted.

**Creating of GUI**

GUI provides a user-friendly and an interactive tool for finding out which Brodmann area has been affected by tumor and the anatomical description of the tumor regions so that it becomes easier for the doctor to use appropriate treatment methodology for the patient.

The GUI provides the user a wide variety of options which include: -

- Loading the MRI images stored anywhere in the computer.
- Viewing each slice of the MRI image.
- Tumor segmentation for each slice of MRI image using fuzzy-c- means method.
- Determining the best segmented tumor among the 5 segmented tumor images.
- Displaying the Brodmann areas and the anatomical description of the tumor region for both right and left hemisphere separately.

## 3 Results

### 3.1 Data Used

Data was acquired for 15 patients using a 3T PHILIPS MRI machine. The acquired data had varying contrasts in ethnicity, age (25-50), tumor locations and sizes. All the images were normalized with 10mm thickness slice gap.

### 3.2 Tumor Segmentation

The user inputs the respective normalized *nifti* file in the GUI interface. Then the slice for which the segmented tumor wants to be seen is selected. As the number of clusters taken in Fuzzy C Means Algorithm is 5, hence 5 different segmented clusters are obtained as shown in figure 3. Then the most appropriate of the five segmented images was selected.

The tumor was delineated from the MR image. The tumor segmentation was found to be effective.

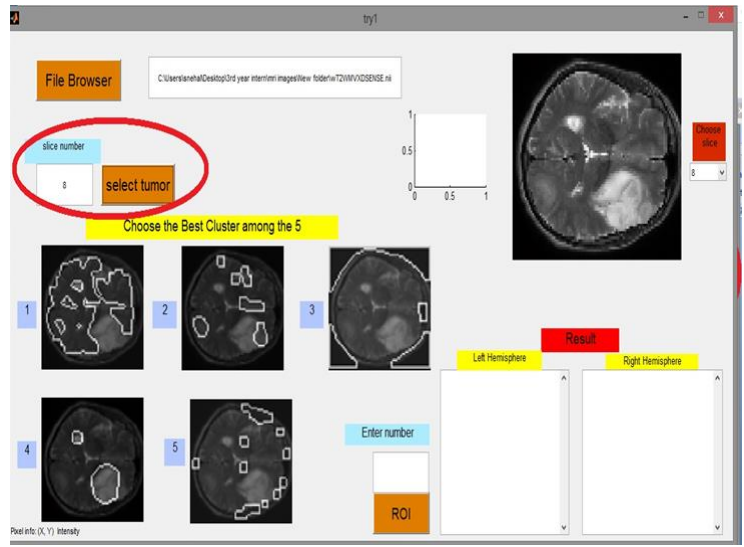

**Fig. 3.** Fuzzy gives 5 output clusters

### 3.3 Brodmann Area Description of the tumor

This step takes in the best selected tumor segmented image from the previous step and tells us the Brodmann areas and anatomical regions of the brain where the tumor exists as shown in figure 4.

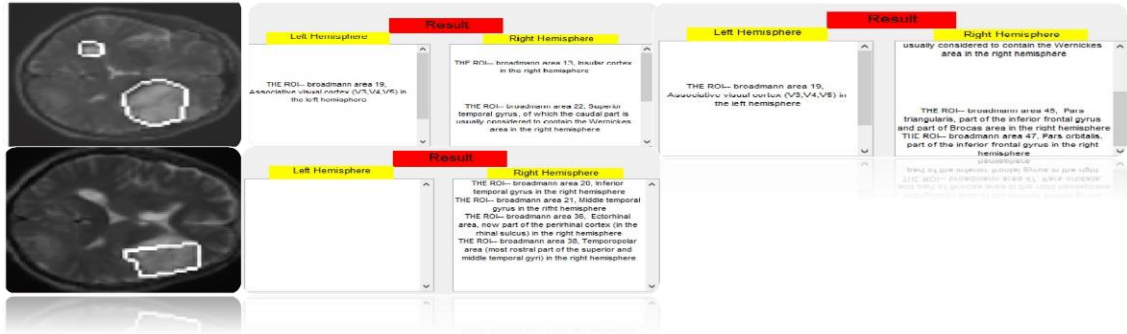

**Fig. 4.** Segmentation and Brodmann Area Detection

### 3.4 Discussion

Automatic Tumor Detection has gained importance as various research groups are currently developing methodologies for the same. This GUI can aid the doctors in correctly and accurately diagnosing tumor. Cobzas et al. proposed a new method where they defined a set of multidimensional features based on templates and atlases and then used them to segment tumor. An argument that region based segmentation methods based on texture features are not suitable for segmentation as they do not differentiate between tumor and patient effectively was presented [10]. Joshi et al. proposed a 2 step process for tumor segmentation. In the first stage they used the gabor filter to extract the features and then contour based methods are used to detect the different objects followed by the discrete level sets for segmentation. They then found their results to be better than the existing methods [11]. Khotanlou et al. used the concept that the brain has to be symmetric about the central plane and hence presence of tumor will disturb the symmetry. This is followed by Fuzzy C- means classification (FCM) and Probabilistic Fuzzy C Means Classification which segments the tumor from the MR image. Testing was done on a dataset of 20 patients having T1 weighted MR images and decent segmentation accuracy was reported by them [12].Manasa et al. used Canny Edge Detection algorithms to segment tumor. They argued Canny Edge Detection is best way to find all the contours irrespective of the edges of white and grey matter as compared to other methods [13]. Zhan et. al made use of multiple modalities in MRI to detect features of tumor which is accompanied by the use of Multiple Classifier system for calculating the brain tumor probability of the image [14]. Xie et al. used the region and boundary information in a hybrid level set(HLS) algorithm to segment the MRI image. The author used 256 axial tumor MR images of 10 patients for tumor segmentation. The method was tested against the manual segmentation method and a mean accuracy of 85.67% was achieved [15].

Tiede et al. described the anatomy of brain in two levels. First level is the volume level which contains info information about the voxel intensity and its relation to the basic regions Second level is the knowledge level which describes the relation between different basic regions in First level. The authors found their method to be superior that the other existing methods [16]. Lancaster et al. proposed a method to determine brain labels from the Talairach atlas. 160 regions were marked in the atlas using volume and sub-volume components. The atlas divided the brain into 5 levels with each structure being divided into substructures in the next level [17]. Mazziotta et al. explained the need of a probabilistic atlas of human brain stating the fact that a probabilistic atlas is tolerant to changes and will preserve variation in brain anatomy. The accuracy of such atlas in clearly labelling the areas improves as the data is increased. This map allowed the user to see the anatomy of the brain and also provided them with a database of each and every region they asked about [18]. MacDonald et al. proposed deformable models for cortex identification. For this purpose, they have used the anatomic segmentation using Proximities (ASP) Algorithm. The advantage of ASP in segmenting human cortex is that it uses simple geo- metric constraints based on anatomical knowledge [19]. Christensen et. al. used a transformation function to accommodate the differences of the brain shape with respect to the standard brain images dictionary. Grenander's shape models were used by them to represent the structure and variation of brain [20]. Van Lemeput et. al. proposed the idea of using a digital brain atlas for prior location of tissues and used MRF (Markov Random Field) to classify brain and non-brain tissues [21].

In this paper, a widely varied dataset of 15 patients was used. T2 weighted sequences were analyzed. Each data was normalized to ensure symmetry between different images of different patients. Fuzzy C Means and K Means Clustering were analyzed for all patients. It was found that Fuzzy C Means was found to give better segmentation accuracy. This method is better as compared to other existing state of the art methods as no training of the data or template for comparison is required. This method takes care of all types of test cases which sometimes are not properly segmented by training methods like Convolutional Neural Network(CNN).

The segmented tumor image is then sent for Brodmann area detection. A database of the Brodmann and anatomical areas of the brain was created and then mapping between the segmented tumor image and the database was done. This method is better as compared to other existing methods as the database created was very vast taking care of all possible Brodmann Areas.

The GUI provides the doctor an easy, user friendly tool to aid them in tumor detection. It also provides the doctor the exact areas where the tumor exists and thus helps them to plan the exact course of treatment for the patient.

## 4      Conclusion

The algorithm is comprehensive in segmenting as well as detecting the Brodmann areas of the tumor. This was tested on a dataset of 15 very widely variant test patients and very good results were achieved. The creation of a GUI makes the whole process very user friendly.